\documentclass{article}
\usepackage{ijcai19}

\usepackage{times}
\usepackage{soul}
\usepackage{url}
\usepackage[hidelinks]{hyperref}
\usepackage[utf8]{inputenc}
\usepackage[small]{caption}
\usepackage{graphicx}
\usepackage{subfigure}
\usepackage{amsmath}
\usepackage{booktabs}
\usepackage{algorithm}
\usepackage{algorithmic}
\usepackage{color}
\usepackage{amsfonts}
\title{Learning Image-Specific Attributes by Hyperbolic Neighborhood Graph Propagation}

\author{
Xiaofeng Xu$^{1,2}$\and
Ivor W. Tsang$^2$\and
Xiaofeng Cao$^2$\and
Ruiheng Zhang$^{2,3}$\And
Chuancai Liu$^{1,4}$\footnote{Corresponding Author}
\affiliations
$^1$School of Computer Science and Engineering, Nanjing University of Science and Technology\\
$^2$Faculty of Engineering and Information Technology, University of Technology Sydney\\
$^3$School of Mechatronical Engineering, Beijing Institute of Technology\\
$^4$Collaborative Innovation Center of IoT Technology and Intelligent Systems, Minjiang University\\
\emails
\{csxuxiaofeng, chuancailiu\}@njust.edu.cn, ivor.tsang@uts.edu.au,\\
\{xiaofeng.cao, ruiheng.zhang\}@student.uts.edu.au\\
\quad
}

\begin{document}

\maketitle

\begin{abstract}
As a kind of semantic representation of visual object descriptions, attributes are widely used in various computer vision tasks. In most of existing attribute-based research, class-specific attributes (CSA), which are class-level annotations, are usually adopted due to its low annotation cost for each class instead of each individual image. However, class-specific attributes are usually noisy because of annotation errors and diversity of individual images. Therefore, it is desirable to obtain image-specific attributes (ISA), which are image-level annotations, from the original class-specific attributes. In this paper, we propose to learn image-specific attributes by graph-based attribute propagation. Considering the intrinsic property of hyperbolic geometry that its distance expands exponentially, hyperbolic neighborhood graph (HNG) is constructed to characterize the relationship between samples. Based on HNG, we define neighborhood consistency for each sample to identify inconsistent samples. Subsequently, inconsistent samples are refined based on their neighbors in HNG. Extensive experiments on five benchmark datasets demonstrate the significant superiority of the learned image-specific attributes over the original class-specific attributes in the zero-shot object classification task.
\end{abstract}

\section{Introduction}
With the rapid development of machine learning techniques and explosive growth in big data, computer vision has made tremendous progress in recent years \cite{lecun2015deep}. As a kind of semantic information, visual attributes received a lot of attention and have been used in various computer vision tasks, such as zero-shot classification (ZSC) \cite{lampert2014attribute} and action recognition \cite{roy2018unsupervised}. 

Attributes are typical nameable properties for describing objects \cite{farhadi2009describing}. They can be the color or shape, or they can be a certain part or a manual description of objects. For example, object \textit{elephant} has attributes \textit{big} and \textit{long nose}, object \textit{zebra} has attributes \textit{striped} and \textit{tail}. Generally, attributes can be detected by recognition systems, and they can be understood by humans. In computer vision tasks, attributes are shared by multiple objects, and thus they can transfer semantic information between different classes \cite{lampert2014attribute}. The essential of attribute-based learning model is that attributes are introduced as an intermediate layer to cascade the feature layer and the label layer.

\begin{figure}[t] 
	\begin{center}
		\includegraphics[width = 0.4\textwidth]{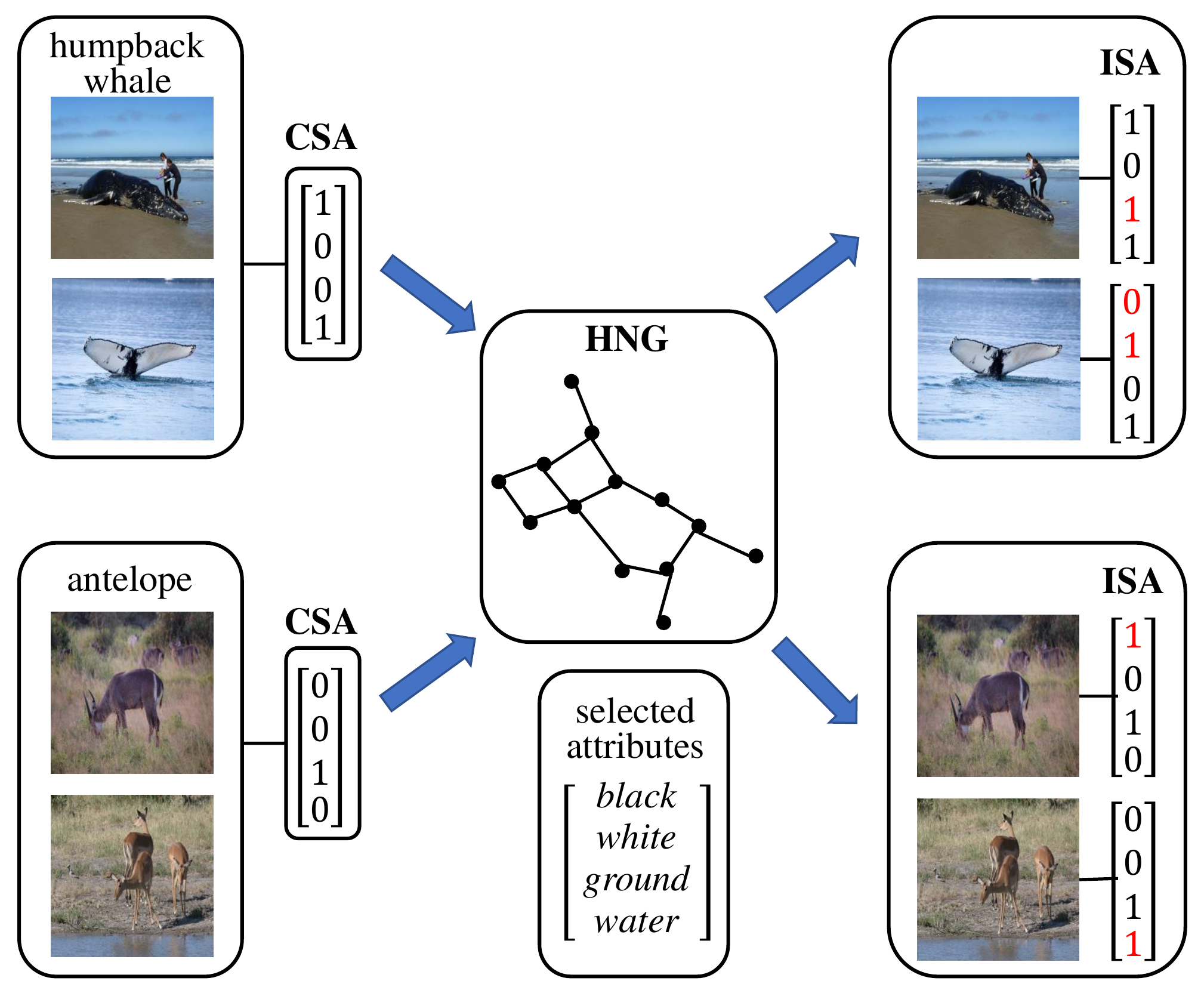}  
	\end{center}
	\caption{Illustration of the proposed image-specific attribute learning model. The left is the original weakly-supervised class-specific attributes (CSA), the middle is the hyperbolic neighborhood graph (HNG), the right is the learned image-specific attributes (ISA).}
	\label{figure_wsisal}
\end{figure}

However, in most of recent attribute-based research, the attribute representation is weakly-supervised in two aspects \cite{akata2016label}. One is the inexact supervision \cite{zhou2017brief}. Existing attribute representation is class-specific (i.e. each class has an attribute representation) due to the high cost of manual annotation for each image. Therefore, class-specific attributes (CSA) are not exact for all the images because of the individual diversity. The other is the inaccurate supervision \cite{zhou2017brief}. Attributes contain noise because of the attribute collecting process.

Supervised learning has achieved great success by learning from high quality labeled data with strong supervision information \cite{zhou2017brief}. Motivated from this, in this paper, we propose to learn image-specific attributes (ISA) from the original weakly-supervised class-specific attributes by hyperbolic neighborhood graph (HNG) propagation. To emphasize similar samples and suppress irrelevant samples, hyperbolic distance is adopted to construct the graph, owing to its intrinsic property of exponentially expansion relative to the Euclidean distance. In the constructed HNG, neighbor samples should have the same attributes, and a sample is subject to caution when its attribute representation is different from that of its neighbors. Therefore, we define neighborhood consistency for each sample to identify the inconsistent samples. After that, inconsistent samples will be refined depending on the attributes of their neighbors. Based on the HNG and neighborhood consistency, image-specific attributes for each sample can be learned from the original weakly-supervised class-specific attributes, and subsequently be leveraged in various attribute-based vision tasks.

Fig. \ref{figure_wsisal} illustrates the proposed image-specific attribute learning model. Image-specific attributes are learned from the original class-specific attributes by hyperbolic neighborhood graph. In Fig. \ref{figure_wsisal}, red attributes in ISA indicate the refinement of corresponding attributes in CSA. It is obvious that the learned ISA is more accurate than the original CSA for the displayed images.

The main contributions of this paper can be summarized as follows:

(1) We propose an image-specific attribute learning  model to learn image-specific attributes from the original weakly-supervised class-specific attributes.

(2) Considering the advantages of hyperbolic geometry and relative neighborhood graph, we design the hyperbolic neighborhood graph to characterize the relationship between samples for subsequent attribute identification and refinement.

(3) We apply the learned image-specific attributes to zero-shot classification task. Experimental results demonstrate the superiority of the learned ISA over the original CSA.

\section{Related Works}
\subsection{Attribute Representation}
Attributes, as a kind of popular visual semantic representation, can be the appearance, a component or the property of objects \cite{farhadi2009describing}. Attributes have been used as an intermediate representation to share information between objects in various computer vision tasks. In zero-shot learning fields, Lampert et al. \cite{lampert2014attribute} proposed the direct attribute prediction model, which categorizes zero-shot visual objects by using attribute-label relationship as the assistant information in a probability prediction model. Akata et al. \cite{akata2016label} proposed the attribute label embedding model, which learns a compatibility function from image features to attribute-based label embeddings. In other fields, Parikh et al. \cite{parikh2011relative} proposed relative attributes to describe objects. Jang et al. \cite{jang2018facial} presented a recurrent learning-based facial attribute recognition model that mimics human observers' visual fixation.

Attributes can be annotated by expert systems or crowdsourcing. Most of the existing attributes are class-specific due to the high cost of annotating per sample. When assigning class-specific attribute representation to samples, attribute noise will be unavoidable because of the individual diversity. In this case, we propose to learn the exact image-specific attributes from the original weakly-supervised class-specific attributes to enhance the semantic representation.

\subsection{Learning from Weakly-Supervised Data}
For most vision tasks, it is easy to get a huge amount of weakly-supervised data, whereas only a small percentage of data can be annotated due to the human cost \cite{zhou2017brief}. In attribute-based tasks, class-specific attributes are weakly-supervised when assigning class-level attributes to samples. To exploit weakly-supervised data, Guar et al. \cite{gaur2017weakly} proposed to extract the implicit semantic object part knowledge by the means of utilizing feature neighborhood context. Mahajan et al. \cite{mahajan2018exploring} presented a unique study of transfer learning to predict hashtags on billions of weakly supervised images. Some weakly-supervised label propagation algorithms are improved by introducing specific constraints (\cite{zhang2018robust,kakaletsis2018label}), adopting the kernel strategy (\cite{zhang2018kernel}), or considering additional information from negative labels (\cite{zoidi2018positive}). In this work, we propose to design the hyperbolic neighborhood graph to learn supervised information from the weakly-supervised class-specific attributes.

\subsection{Hyperbolic Geometry}
Hyperbolic geometry is a non-Euclidean geometry with a constant negative Gaussian curvature. A significant intrinsic property in hyperbolic geometry is the exponential growth, instead of the polynomial growth as in euclidean geometry. As illustrated in Figure \ref{figure_fig1_a}, the number of intersections within a distance of $ d $ from the center rises exponentially, and the line segments have equal hyperbolic length. Therefore, hyperbolic geometry outperforms euclidean geometry in some special tasks, such as learning hierarchical embedding \cite{ganea2018hyperbolic}, question answer retrieval \cite{tay2018hyperbolic} and hyperbolic neural networks \cite{nitta2018hyperbolic}. In this work, we adopt the hyperbolic distance to construct the graph, and consequently, similar samples will be emphasized and irrelevant samples will be suppressed.

\section{Approach}
In this section, we present the formulation of learning image-specific attributes by a tailor-made hyperbolic neighborhood graph.

\subsection{Notations}
Let $ N $ be the number of samples in dataset $ \mathcal{S} =\{(x_i,y_i)\}^N_{i=1} $, which is also the number of vertices in the graph. $ L $ is the number of the labels in $ \mathcal{S} $.  $ \mathbf{A} \in \{0,1\}^{M \times L} $ is the initial class-specific attribute representation for each label, where $M$ denotes the dimension of attribute representation. $ \mathbf{\tilde{A}} \in \{0,1\}^{M \times N} $ is the learned image-specific attribute representation for each sample. 

\subsection{Image-Specific Attribute Learning}
In this work, we take a graph-based approach to learn the image-specific attributes. The proposed approach is conducted in four steps: 1) calculating hyperbolic distance between samples; 2) constructing hyperbolic neighborhood graph; 3) identifying inconsistent samples; 4) refining the attributes of inconsistent samples.

\subsubsection{Calculation of Hyperbolic Distance}
We construct a graph to characterize the relationship between samples. Due to the negative curvature of hyperbolic geometry, hyperbolic distance between points increases exponentially relative to euclidean distance. In hyperbolic geometry, the distance between irrelevant samples will be exponentially greater than the distance between similar samples. Therefore, hyperbolic distance is adopted to construct the graph. The relationship between samples represented in the hyperbolic-based graph can emphasize similar samples and suppress irrelevant samples.

Poincar\'{e} disk \cite{nickel2017poincare}, as a representative model of hyperbolic geometry, is employed to calculate the hyperbolic distance. Poincar\'{e} disk model $(\mathcal{B}^n,g_p)$ is defined by the manifold $\mathcal{B}^n=\{x\in \mathbb{R}^n:\|x\|<1\}$ equipped with the following Riemannian metric:
\begin{equation}
	g_p(x) = \left(\frac{2}{1-\|x\|^2}\right)^2g_e,
	\label{eq_poincare}
\end{equation}
where $g_e$ is the Euclidean metric tensor with components $\mathbf{I}_n$ of $\mathbb{R}^n$.

The induced hyperbolic distance between two samples $x_i,x_j \in \mathcal{B}^n$ is given by:
\vskip -0.15in
\begin{small}
	\begin{equation}
	d(x_i,x_j) = \mathrm{cosh}^{-1}\left(1+2\frac{\|x_i-x_j\|^2}{(1-\|x_i\|^2)\cdot(1-\|x_j\|^2)}\right)
	\label{eq_hypdist} 
	\end{equation}
\end{small}

\subsubsection{Hyperbolic Neighborhood Graph}
To measure the proximity of samples in dataset $ \mathcal{S} $, we design a hyperbolic neighborhood graph, which is tailor-made based on the relative neighborhood graph \cite{toussaint1980relative} by introducing hyperbolic geometry. 

Let $G=(V,E)$ be a hyperbolic neighborhood graph with vertices $v\in V$, the set of samples in $ \mathcal{S} $, and edges $(v_i,v_j)\in E$ corresponding to pairs of neighboring vertices $v_i$ and $ v_j $. To evaluate whether two vertices  $v_i$ and $ v_j $ are neighbors of each other, we adopt the definition of ``relatively close'' \cite{lankford1969regionalization}. The pair $ v_i $ and $ v_j $ are neighbors if:
\begin{equation}
	d(v_i,v_j) \leq \mathrm{max}(d(v_i,v_k),d(v_j,v_k)) \quad \forall k \neq i,j
	\label{eq_graphdist} 
\end{equation}
where $ d(\cdot,\cdot) $ is the hyperbolic distance defined in \eqref{eq_hypdist}.

\begin{figure}[t]
	\centering 
	\subfigure[Poincar\'{e} disk\protect\footnotemark]{ 
		\label{figure_fig1_a} 
		\includegraphics[width=0.2\textwidth]{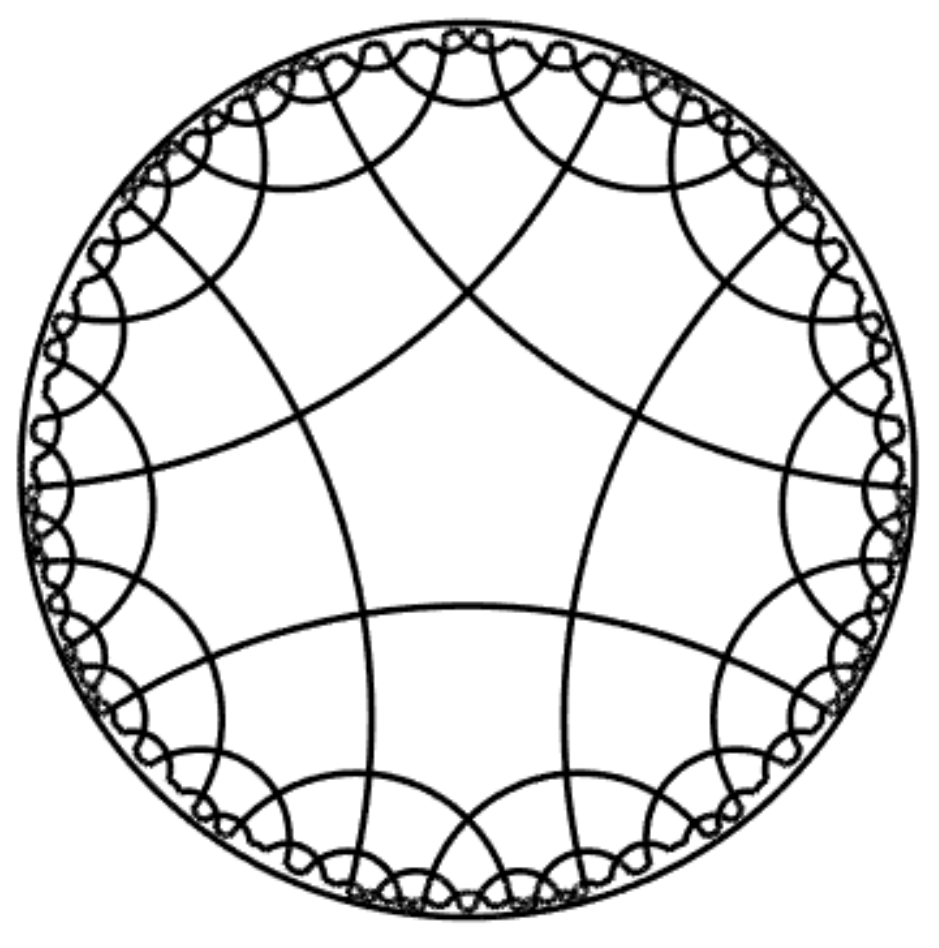} 
	} 
	\subfigure[HNG]{ 
		\label{figure_fig1_b} 
		\includegraphics[width=0.2\textwidth]{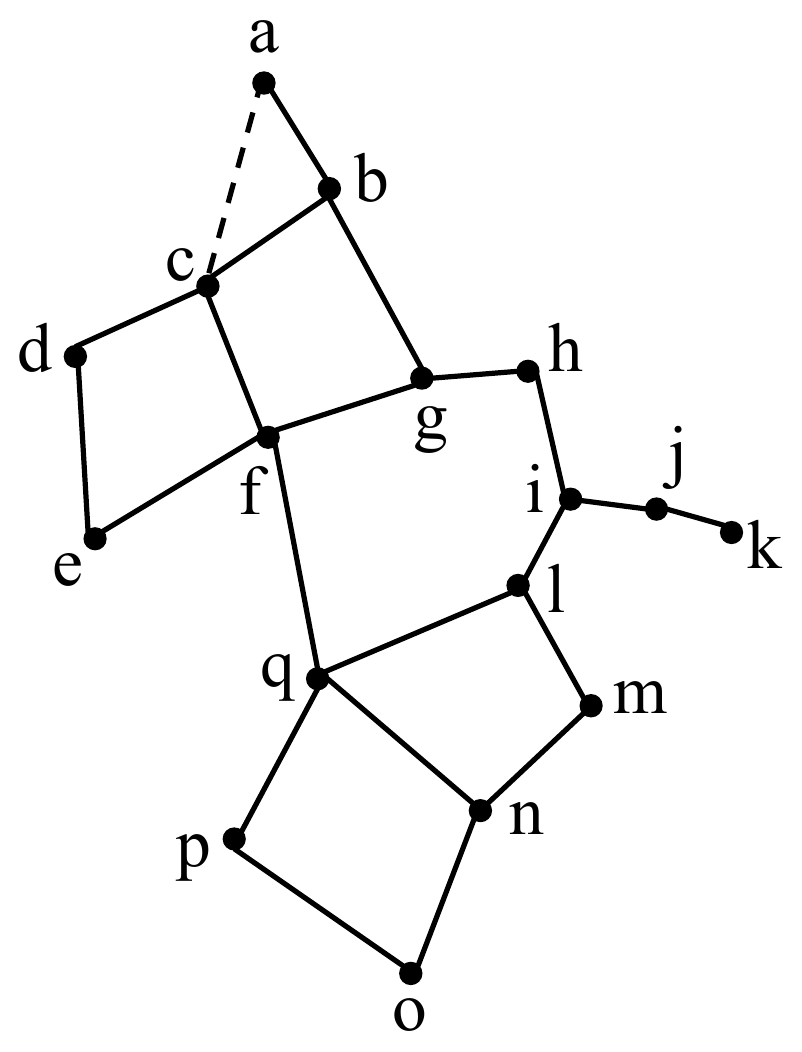} 
	} 
	\caption{(a) Pentagonal tiling of the Poincar\'{e} disk hyperbolic geometry, all black line segments have equal hyperbolic length. (b) Illustration of hyperbolic neighborhood graph (HNG) constructed from a set of points.} 
	\label{figure_fig1} 
\end{figure}

Intuitively, two vertices $v_i$ and $ v_j $ are neighbors if, and only if, in the triangle composed of vertices $v_i$, $ v_j $ and any other vertex $v_k$, the side between $v_i$ and $ v_j $ is not the longest side. Figure \ref{figure_fig1_b} illustrates the HNG constructed from a set of points. The solid edge indicates that the end vertices of the edge are neighbors, and the dotted edge indicates that the end vertices are not neighbors. For example, the vertices $a$ and $b$ are neighbors, while the vertices $a$ and $c$ are not neighbors because the edge $(a,c)$ is longer than both edges $(a,b)$ and $(c,b)$.

\subsubsection{Identification of Inconsistent Samples}
\footnotetext{https://www.mathworks.com/matlabcentral/fileexchange/34417-pentagonal-tiling-of-the-poincare-disc}

After constructing the hyperbolic neighborhood graph, we define neighborhood consistency as the confidence degree for each vertex.
Neighborhood consistency is the similarity between the attribute value of one sample and that of expected attribute value from its neighbor samples.
In HNG, samples should be neighborhood consistent, i.e., each sample has the similar attributes to its neighbors. Therefore, we can calculate the neighborhood consistency of each sample based on it and its neighbors' attribute values.

Given the hyperbolic distance assigned to each edge, we first calculate the weights of all the associated edges for each vertex. Suppose that vertex $v$ has $k$ neighbor vertices which are associated with $k$ edges $\{e_i\}^k_{i=1}$, and the hyperbolic distances corresponding to the edges are $\{h_i\}^k_{i=1}$. The relative weights of each edge for vertex $v$ are calculated based on Inverse Distance Weighting \cite{shepard1968two} as follows:
\begin{equation}
	w_i = \frac{h_i^{-p}}{\sum_{j=1}^{k} h_j^{-p}},
	\label{eq_weight}
\end{equation}
where $p$ is an arbitrary positive real number (usually sets to 1 or 2).
For each vertex, the weight $w_i$ of the associated edge $e_i$ is inversely related to its distance $h_i$, and the sum of the weights of all associated edges is equal to $1$. 

Each vertex in HNG represents a sample derived from dataset $ \mathcal{S} $.
For vertex $ v $, suppose its corresponding sample is $(x_v,y_v)$ with the attribute representation ${a}_v$. Assume that in the graph, vertex $v$ has $k$ neighbor vertices $\{v_i\}^k_{i=1}$ with the attribute representation $\{{a}_{v_i}\}^k_{i=1}$, and it has $k$ associated edges $\{e_i\}^k_{i=1}$ with the weights $\{w_i\}^k_{i=1}$. Then, the neighborhood consistency of vertex $v$  can be calculated as follows:
\begin{equation}
	J(v) = z_v a_v+(1-z_v)(1-a_v),
	\label{eq_conf}
\end{equation}
where $ a_v\in\{0,1\} $ is the attribute value of vertex $v$. $z_v$ is the expected attribute value of $v$, which is calculated based on the attributes of its neighbor vertices: 
\begin{equation}
	z_v = \sum_{i=1}^{k}w_i a_{v_i},
	\label{eq_confz}
\end{equation}
where $ \sum_{i=1}^{k}w_i =1 $.

After calculating the neighborhood consistency of vertex $v$, we can identify whether the corresponding sample $ x_v $ is neighborhood consistent based on the neighborhood consistency $ J(v) $ as follows:
\begin{equation}
	\left\{\begin{aligned}
	&J(v) \geq \theta, \text{$ x_v $ is neighborhood consistent}, \\ 
	&J(v) < \theta, \text{$ x_v $ is inconsistent},
	\end{aligned}\right.
	\label{eq_ident}
\end{equation}
where $\theta$ is a hyperparameter that constraints the consistency degree when identifying inconsistent samples. 

\subsubsection{Refinement of Attributes for Inconsistent Samples}
After identifying inconsistent samples, we can learn the image-specific attributes by refining the initial weakly-supervised class-specific attributes. We keep the attributes of consistent samples and refine the attributes of inconsistent samples. Considering that the initial class-specific attribute $a\in\{0,1\}$, we directly alter the attribute value from 0 to 1 or from 1 to 0 for inconsistent samples.

Our formalization allows associating keeping consistent samples and refining inconsistent examples when playing on the hyperparameter $\theta$. There are two errors in the processing of identification and refinement. One is the false positive (FP) error that consistent samples are identified as inconsistent samples by mistake, and the other is the false negative (FN) error that inconsistent samples can not be identified. The closer $\theta$ is to $0$, the fewer inconsistent samples can be identified, and thus the FP error will decrease and the FN error will increase. The closer $\theta$ is to $1$, the more consistent samples will be identified to be inconsistent by mistake, and thus the FP error will increase and the FN error will decrease. Therefore, we can constrain $\theta$ to balance these two errors.

The algorithm of image-specific attribute learning model is given in Algorithm \ref{alg_alg1}.
\begin{algorithm}[h]
	\caption{Image-Specific Attribute Learning Model}
	\begin{algorithmic}[1]
		\REQUIRE ~~\\
		Image features; \\
		Class-specific attributes $\mathbf{A} \!= \! \{\mathbf{a}_i\}^L_{i=1}$ for each label. \\
		\ENSURE ~~\\
		Image-specific attributes $\mathbf{\tilde{A}} \!=\! \{\mathbf{\tilde{a}}_i\}^{N}_{i=1}$ for each sample. \\ 
		\vspace{2mm}
		\textbf{// Construct Hyperbolic Neighborhood Graph}
		\STATE Calculate the hyperbolic distance between each sample pair by Eq.\eqref{eq_hypdist}; \\
		\STATE Construct hyperbolic neighborhood graph by Eq.\eqref{eq_graphdist};\\
		\textbf{// Identify Inconsistent Samples}
		\STATE Calculate weights of the edges in HNG by Eq.\eqref{eq_weight};
		\STATE Calculate the neighborhood consistency of each vertex in HNG by Eq.\eqref{eq_conf};
		\STATE Identify inconsistent examples by Eq.\eqref{eq_ident}; \\
		\textbf{// Refine Attributes for Inconsistent Samples}
		\STATE Refine attributes of the inconsistent samples.
	\end{algorithmic} 
	\label{alg_alg1}	
\end{algorithm}

\subsection{ISA for Zero-Shot Classification}
After obtaining the image-specific attribute representation $ \mathbf{\tilde{A}} \! \in\! \{0,1\}^{M \times N} $ for training data $ \mathcal{S}\! =\!\{(x_i,y_i)\}^N_{i=1} $, we can easily apply it to existing zero-shot classification methods. We learn a function  $ f\!:\!\mathbf{X}\!\rightarrow\! \mathbf{\tilde{A}} $ which maps image features to attribute embeddings by optimizing following empirical risk:
\begin{equation}
\mathop{\min}_{\mathbf{W}} L(f(\mathbf{X};\mathbf{W}) , \mathbf{\tilde{A}}) +\Omega (\mathbf{W}),
\label{eq_zsltr}
\end{equation}
where $L()$ is the loss function and $\Omega()$ is the regularization term. $ \mathbf{W} \in R^{M \times d} $ is the parameter of mapping function defined by various specific zero-shot classification methods \cite{xian2018zero}.

At test stage, image features of test samples are mapped to attribute embedding space via the learned function $ f $. Given the attribute representation $\{\mathbf{a}_i\}^K_{i=1}$ of test labels ($ \#K $), test sample $x$ is classified as follows:
\begin{equation}
y = arg\mathop{\min}_{i} dist(f(x;\mathbf{W}),\mathbf{a}_i)
\label{eq_zslts}
\end{equation}
where $dist(\cdot,\cdot)$ denotes the cosine distance metric.

\subsection{Complexity}
In this work, all the attributes are handled using the same graph constructed based on the image features. Therefore, the complexity of the attribute learning algorithm is $ \mathcal{O}(G) $, i.e. the complexity of constructing the HNG. When constructing the graph, we first calculate the distance between all pairs of samples, which requires $ \mathcal{O}(N^2) $ operations. Then we filter the edges based on \eqref{eq_graphdist}, which requires $ \mathcal{O}(N) $ operations. Thus, the complexity of constructing the graph is $ \mathcal{O}(N^3) $.

\section{Experiments}
To evaluate the effectiveness of learned image-specific attributes, we conduct experiments of zero-shot object classification task on five benchmark datasets.

\subsection{Settings}
\noindent\textbf{Datasets.} Experiments are conducted on five zero-shot classification benchmark datasets: (1) Animal with Attribute (AwA)~\cite{lampert2014attribute}, (2) Animal with Attribute 2 (AwA2)~\cite{xian2018zero}, (3) attribute-Pascal-Yahoo (aPY)~\cite{farhadi2009describing}, (4) Caltech-UCSD Bird 200-2011 (CUB)~\cite{welinder2011caltech}, and (5) SUN Attribute Database (SUN) \cite{patterson2012sun}. ResNet-101~\cite{he2016deep} is used to extract deep features for experiments. We use 85, 85, 64, 312 and 102-dimensional attribute representation for AwA, AwA2, aPY, CUB and SUN datasets, respectively. Statistic information of these datasets is summarized in Table \ref{table_datasetstat}.

\begin{table}[t]
	\centering
	\renewcommand\arraystretch{1.1}
	\setlength{\tabcolsep}{2mm}{
		\begin{tabular}{c|c|c|c}
			\hline\hline
			Datasets &Attributes &Classes &Images \\ \hline\hline
			AwA &85 &40 + 10 &19832 + 4958  \\ 
			AwA2 &85 &40 + 10 &23527 + 5882 \\ 
			aPY &64 &20 + 12 &5932 + 1483 \\ 
			CUB &312 &150 + 50 &7057 + 1764 \\
			SUN &102 &645 + 72 &10320 + 2580 \\  \hline	\hline		
	\end{tabular}}
	\caption{Statistics for datasets in terms of number of attributes, number of classes in training + test, number of images in training + test.}
	\label{table_datasetstat}
\end{table}

\noindent\textbf{Evaluation protocol.} 
We follow the experimental settings proposed in \cite{xian2018zero} for all the compared methods to get the fair comparison results. We use the mean class accuracy, i.e. per-class averaged top-1 accuracy, as the criterion of assessment. Mean class accuracy is calculated as follows: 
\begin{equation}
acc=\frac{1}{K} \sum_{y\in\mathcal{Y}_u}\frac{\# \mathrm{correct\; predictions\; in\;} y}{ \# \mathrm{samples\; in\;} y},
\end{equation}
where $K$ is the number of test classes, $\mathcal{Y}_u$ is the set comprised of all the test labels.

\subsection{Compared Methods}
We compare our model to following representative zero-shot classification baselines. (1) Four popular ZSC baselines. DAP \cite{lampert2014attribute} utilizes attribute representation as intermediate layer to cascade visual features and labels. 
ALE \cite{akata2016label} learns the compatibility function between visual features and attribute embeddings. SAE \cite{kodirov2017semantic} learns a low dimensional semantic representation of visual features. MFMR \cite{xu2017matrix} learns a projection based on matrix tri-factorization with manifold regularization. (2) Four latest ZSC baselines. SEZSL \cite{kumar2018generalized} generates novel exemplars from seen and unseen classes to conduct the classification. SPAEN \cite{chen2018zero} tackles the semantic loss problem for the embedding-based zero-shot tasks. LESAE \cite{liu2018zero} learns a low-rank mapping to link visual features and attribute representation. XIAN \cite{xian2018feature} synthesizes CNN features conditioned on class-level semantic information for zero-shot classification.

\begin{table}[t]
	\centering
	\renewcommand\arraystretch{1.1}
	\setlength{\tabcolsep}{2.5mm}{
		\begin{tabular}{c|c|c|c|c|c}
			\hline\hline
			Methods &AwA &AwA2 &aPY &CUB &SUN \\ \hline\hline
			DAP &44.1 &46.1 &33.8 &40.0 &39.9 \\ 
			ALE &59.9 &62.5 &39.7 &54.9 &58.1 \\ 
			SAE &53.0 &54.1 &8.3 &33.3 &40.3 \\ 
			MFMR &64.89 &61.96 &30.77 &51.45 &51.88 \\
			SEZSL &68.2 &66.3 &- &54.1 &61.2 \\
			SPAEN &58.5 &- &24.1 &55.4 &59.2 \\
			LESAE &66.1 &68.4 &40.8 &53.9 &60.0 \\
			XIAN &68.2 &- &- &57.3 &60.8 \\\hline
			Ours &\textbf{71.25} &\textbf{70.64} &\textbf{41.37} &\textbf{58.55} &\textbf{63.59} \\ \hline	\hline	
	\end{tabular}}
	\caption{Zero-shot classification results (in $\%$) on AwA, AwA2, aPY, CUB and SUN datasets. Boldface indicates the best.}
	\label{table_zslall}
\end{table}

\subsection{Comparison with the State-of-the-Art}
To evaluate the proposed image-specific attribute learning model, we compare it to several state-of-the-art ZSL baselines. The zero-shot classification results on five benchmark datasets are shown in Table \ref{table_zslall}. It is obvious that the proposed method achieves the best results on all datasets. Specifically, the accuracies on five datasets (i.e. AwA, AwA2, aPY, CUB and SUN) are increased by $ 3.05\%, 2.24\%, 0.57\%, 1.25\%, 2.39\% $, respectively ($ 1.90\% $ on average) comparing to the strongest competitor. In zero-shot visual object classification tasks, attributes are adopted as the intermediate semantic layer for transferring information between seen classes and unseen classes. Hence, ability of accurately recognizing attributes governs the performance of the entire zero-shot classification system. In this work, the proposed method achieves the state-of-the-art performance, which demonstrates the superiority of the learned image-specific attributes over the original class-specific attributes.

\begin{figure*}[htbp] 
	\begin{center}
		\includegraphics[width = 0.9\textwidth]{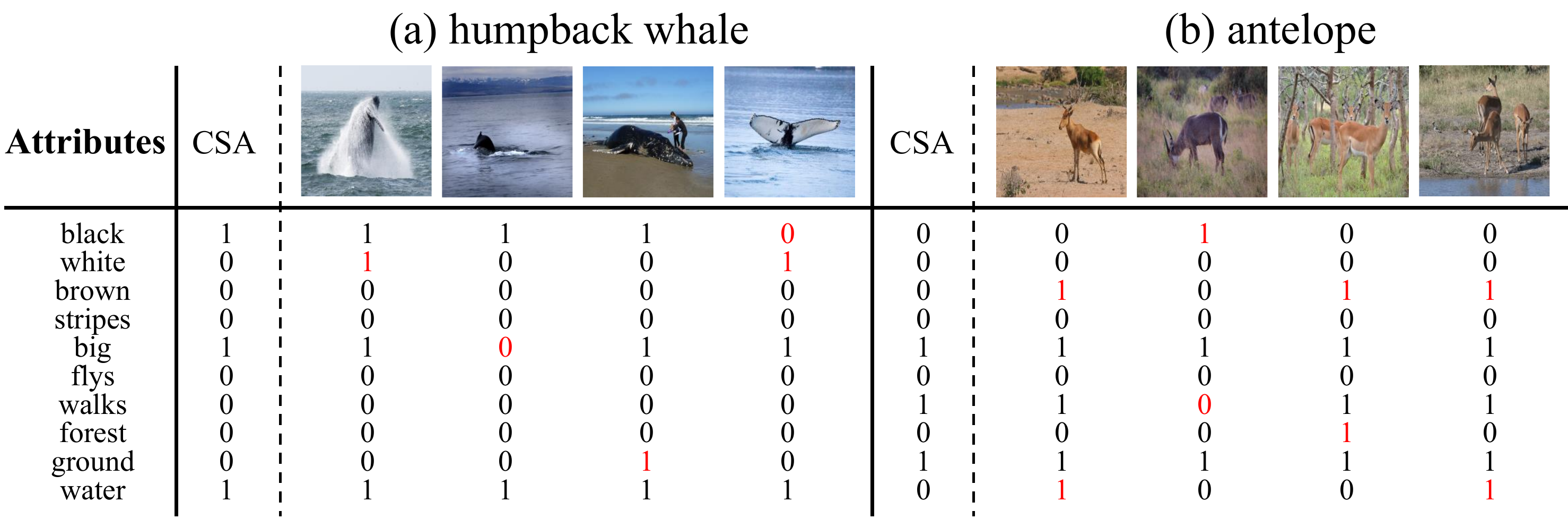}  
	\end{center}
	\caption{Visualization of learned image-specific attributes of several images derived from AwA2 dataset. The first column of both (a) and (b) are class-specific attributes (CSA) of labels. The right four columns are learned image-specific attributes. Numbers in red indicate the refinement of original class-specific attributes.}
	\label{figure_vis}
\end{figure*}

\begin{figure}[t]
	\centering 
	\subfigure[Number of refined attributes]{ 
		\label{figure_theta_a} 
		\includegraphics[width=0.225\textwidth]{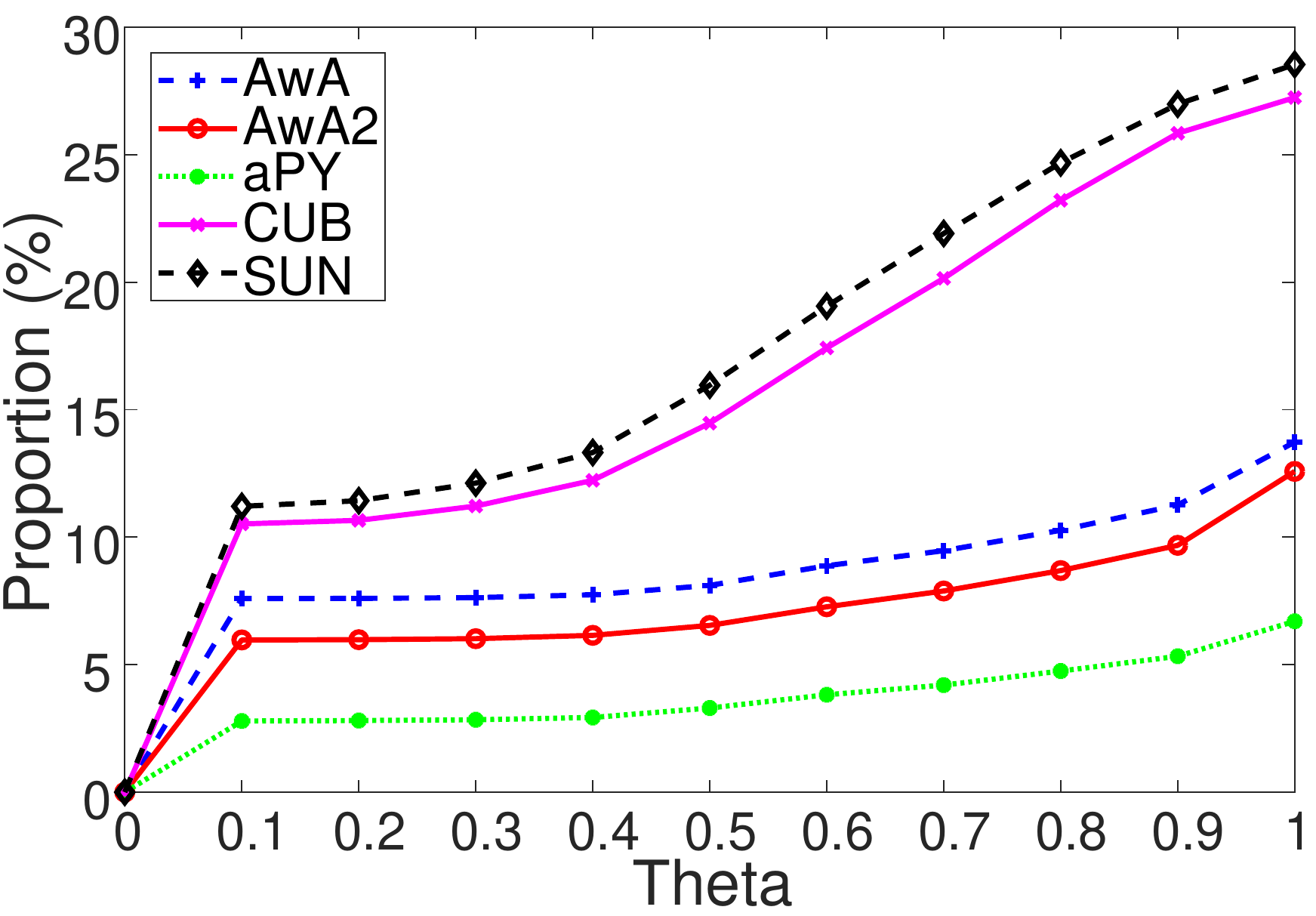} 
	} 
	\subfigure[Classification accuracy]{ 
		\label{figure_theta_b} 
		\includegraphics[width=0.225\textwidth]{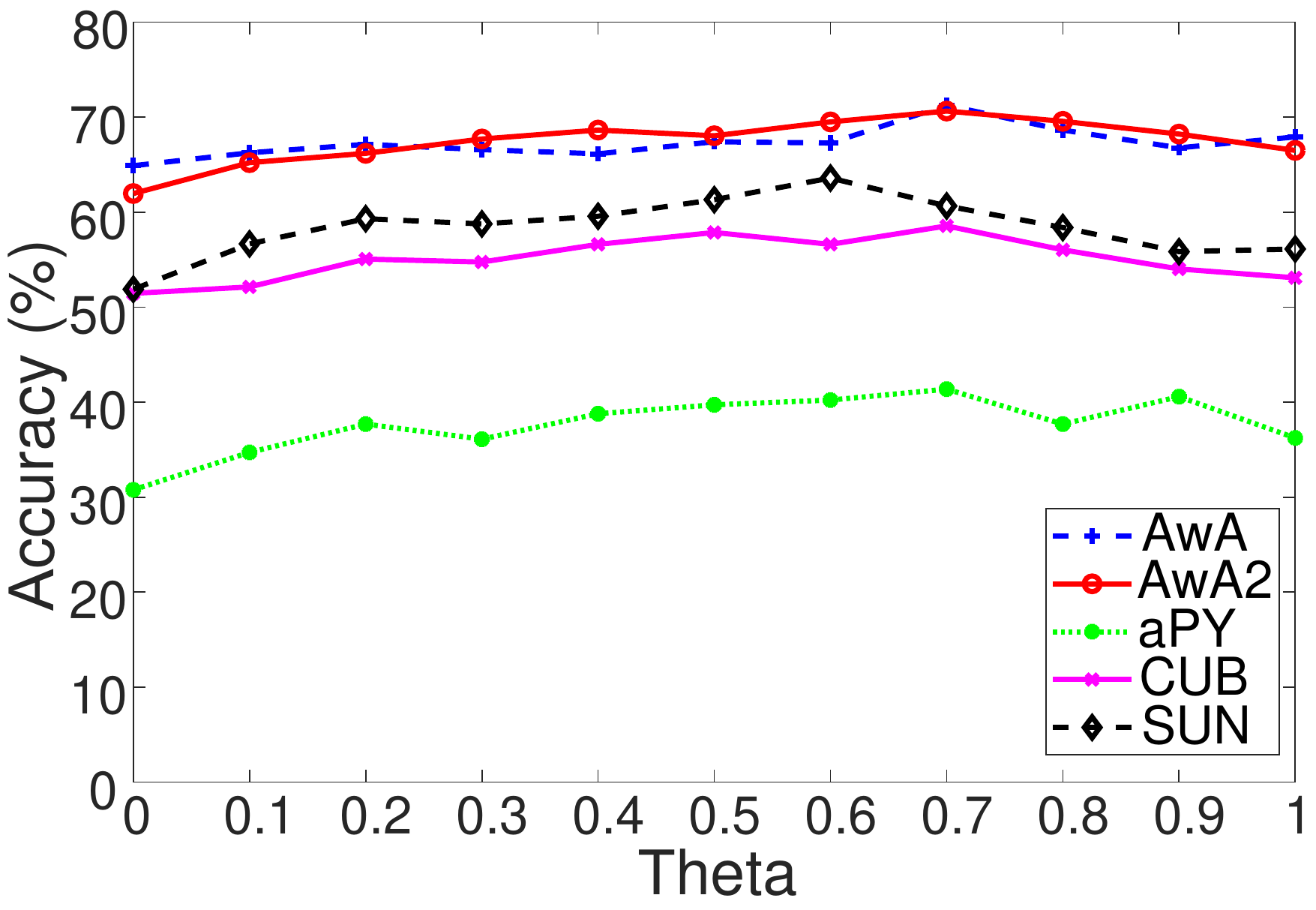} 
	} 
	\caption{Results (in $\%$) influenced by the hyperparameter $\theta$ on five datasets. (a) Porportion of refined attributes; (b) Classification accuracy.} 
	\label{figure_theta} 
\end{figure}

\subsection{Ablation Analysis}
We conduct ablation analysis to further evaluate the effectiveness of the learned image-specific attributes.

\begin{table}[t]
	\centering
	\renewcommand\arraystretch{1.1}
	\setlength{\tabcolsep}{0.8mm}{
		\begin{tabular}{c||c||c|c|c|c|c}
			\hline\hline
			Methods &Graph &AwA &AwA2 &aPY &CUB &SUN \\ \hline\hline
			SAE+CSA &        &53.0 &54.1 &8.3 &33.3 &40.3 \\ 
			SAE+ISA &HCG &54.94 &54.53 &8.33 &37.29 &43.54 \\
			SAE+ISA &ENG &\textbf{57.63} &55.06 &8.33 &35.21 &42.08 \\
			SAE+ISA &HNG &56.44 &\textbf{56.8} &\textbf{8.33} &\textbf{38.01} &\textbf{46.68} \\\hline
			MFMR+CSA &        &64.89 &61.96 &30.77 &51.45 &51.88\\
			MFMR+ISA &HCG &65.17 &62.64 &33.46 &54.34 &56.53 \\
			MFMR+ISA &ENG &69.00 &62.17 &31.38 &54.62 &56.67 \\
			MFMR+ISA &HNG &\textbf{69.49} &\textbf{64.97} &\textbf{35.40} &\textbf{56.61} &\textbf{58.26} \\
			\hline	\hline	
	\end{tabular}}
	\caption{Zero-shot classification results (in $\%$) of using original class-specific attributes (CSA) and image-specific attributes (ISA) learned by hyperbolic complete graph (HCG), euclidean neighborhood graph (ENG) and hyperbolic neighborhood graph(HNG), respectively. Boldface indicates the best.}
	\label{table_zslimp}
\end{table}

\subsubsection{Evaluation on Learned Image-Specific Attributes}
In the first ablation experiment, we evaluate the learned image-specific attributes on two baselines (i.e. SAE and MFMR). We compare the original class-specific attributes with the image-specific attributes learned by hyperbolic neighborhood graph, hyperbolic complete graph (HCG) and euclidean neighborhood graph (ENG), respectively. The comparison results are exhibited in Table \ref{table_zslimp}.

Hyperbolic geometry has a constant negative curvature, where hyperbolic distance increases exponentially relative to euclidean distance. Comparing to the euclidean neighborhood graph, HNG can emphasize the influence of similar samples and suppress irrelevant samples. Therefore, we construct hyperbolic neighborhood graph to characterize samples in the hyperbolic geometry. From Table \ref{table_zslimp}, we can see that ISA leaned by HNG outperforms ENG on all datasets except the AwA dataset for SAE. Since the number of images per class in AwA dataset is much greater than CUB and SUN datasets, most of the neighbor samples in AwA may come from the same class, which have the same attributes. Therefore, fewer samples in AwA would be detected as inconsistent samples, and consequently, the improvement on AwA dataset is not significant comparing to that on CUB and SUN datasets.

Another key component of the proposed model is adopting relative neighborhood graph instead of complete graph to characterize samples. Neighborhood graph is constructed based on the relative relation, which can extract the perceptual relevant structure of data. From Table \ref{table_zslimp}, it is obvious that HNG outperforms HCG on all the datasets, which demonstrates that the relative neighborhood graph has an advantage over the complete graph in describing the proximity between samples.

\subsubsection{Visualization of Learned Image-Specific Attributes}
In the second ablation experiment, we illustrate some learned image-specific attributes of images randomly selected from AwA2 dataset in Fig. \ref{figure_vis}. We compare the learned image-specific attributes with the original class-specific attributes, and numbers in red indicate the refinement of original attributes. Original class-specific attributes are usually noisy due to annotation errors in vision tasks and diversity of individual images. When assigning them to images, the attributes for specific image are inexact because of the individual diversity. For example, normally, \textit{humpback whales} are \textit{black} and live in the \textit{water}. However, when their tails are photographed from below, they are \textit{white} (as shown in the fourth image in Fig. \ref{figure_vis}(a)). And in some images, they may appear on the \textit{ground} (as shown in the third image in Fig. \ref{figure_vis}(a)). From Fig. \ref{figure_vis}, we can see that the learned image-specific attributes are more exact and reliable than the original class-specific attributes. In other words, learned image-specific attributes contain stronger supervision information.

\subsubsection{Parameter Sensitivity Analysis}
In the third ablation experiment, we analyze the influence of the hyperparameter $\theta$. The results are shown in Fig. \ref{figure_theta}. In the proposed model, $\theta$ controls the processing of identifying inconsistent samples. Specifically, the closer $\theta$ is to $1$, the easier samples are identified as neighborhood inconsistent, and thus the false positive error will increase and the false negative error will decrease. From Fig. \ref{figure_theta_b}, we can see that the classification accuracies rise first and then decline as $\theta$ increases from $0$ to $1$. The accuracies reach the peak when $\theta$ is close to $0.7$. In conclusion, $\theta$ balances the false positive error and the false negative error, and consequently influences the quality of the learned image-specific attributes.

\section{Conclusion}
In this paper, a novel hyperbolic neighborhood graph is designed to learn the image-specific attributes. Considering the intrinsic property of hyperbolic geometry that distance increases exponentially, we adopt hyperbolic distance to construct the graph for characterizing samples. After identifying inconsistent samples based on defined neighborhood consistency, we refine the original class-specific attributes to obtain the image-specific attributes. Experimental results demonstrate the advantages of the learned ISA over the original CSA from both accuracy evaluation and visual interpretability in the zero-shot classification task. 
In this work, we assume that attributes are independent of each other and handled separately. In the future, we will consider the interdependence between attributes and handle attributes of inconsistent samples in a unified framework.

\section*{Acknowledgments}
This work was supported in part by NSFC under Grant 61373063 and Grant 61872188, in part by the Project of MIIT under Grant E0310/1112/02-1, in part by the Collaborative Innovation Center of IoT Technology and Intelligent Systems of Minjiang University under Grant IIC1701, in part by ARC under Grant FT130100746, Grant LP150100671 and Grant DP180100106, and in part by China Scholarship Council.

\clearpage
\bibliographystyle{named}
\bibliography{ijcai19}

\begin{thebibliography}{}

\bibitem[\protect\citeauthoryear{Akata \bgroup \em et al.\egroup
  }{2016}]{akata2016label}
Zeynep Akata, Florent Perronnin, Zaid Harchaoui, and Cordelia Schmid.
\newblock Label-embedding for image classification.
\newblock {\em IEEE TPAMI}, 38(7):1425--1438, 2016.

\bibitem[\protect\citeauthoryear{Chen \bgroup \em et al.\egroup
  }{2018}]{chen2018zero}
Long Chen, Hanwang Zhang, Jun Xiao, Wei Liu, and Shih-Fu Chang.
\newblock Zero-shot visual recognition using semantics-preserving adversarial
  embedding networks.
\newblock In {\em CVPR}, 2018.

\bibitem[\protect\citeauthoryear{Farhadi \bgroup \em et al.\egroup
  }{2009}]{farhadi2009describing}
Ali Farhadi, Ian Endres, Derek Hoiem, and David Forsyth.
\newblock Describing objects by their attributes.
\newblock In {\em CVPR}, 2009.

\bibitem[\protect\citeauthoryear{Ganea \bgroup \em et al.\egroup
  }{2018}]{ganea2018hyperbolic}
Octavian-Eugen Ganea, Gary B{\'e}cigneul, and Thomas Hofmann.
\newblock Hyperbolic entailment cones for learning hierarchical embeddings.
\newblock In {\em ICML}, 2018.

\bibitem[\protect\citeauthoryear{Gaur and Manjunath}{2017}]{gaur2017weakly}
Utkarsh Gaur and BS~Manjunath.
\newblock Weakly supervised manifold learning for dense semantic object
  correspondence.
\newblock In {\em ICCV}, 2017.

\bibitem[\protect\citeauthoryear{He \bgroup \em et al.\egroup
  }{2016}]{he2016deep}
Kaiming He, Xiangyu Zhang, Shaoqing Ren, and Jian Sun.
\newblock Deep residual learning for image recognition.
\newblock In {\em CVPR}, 2016.

\bibitem[\protect\citeauthoryear{Jang \bgroup \em et al.\egroup
  }{2018}]{jang2018facial}
Jinhyeok Jang, Hyunjoong Cho, Jaehong Kim, Jaeyeon Lee, and Seungjoon Yang.
\newblock Facial attribute recognition by recurrent learning with visual
  fixation.
\newblock {\em IEEE Trans. Cybern.}, 2018.

\bibitem[\protect\citeauthoryear{Kakaletsis \bgroup \em et al.\egroup
  }{2018}]{kakaletsis2018label}
Efstratios Kakaletsis, Olga Zoidi, Ioannis Tsingalis, Anastasios Tefas, Nikos
  Nikolaidis, and Ioannis Pitas.
\newblock Label propagation on facial images using similarity and dissimilarity
  labelling constraints.
\newblock In {\em ICIP}, 2018.

\bibitem[\protect\citeauthoryear{Kodirov \bgroup \em et al.\egroup
  }{2017}]{kodirov2017semantic}
Elyor Kodirov, Tao Xiang, and Shaogang Gong.
\newblock Semantic autoencoder for zero-shot learning.
\newblock In {\em CVPR}, 2017.

\bibitem[\protect\citeauthoryear{Kumar~Verma \bgroup \em et al.\egroup
  }{2018}]{kumar2018generalized}
Vinay Kumar~Verma, Gundeep Arora, Ashish Mishra, and Piyush Rai.
\newblock Generalized zero-shot learning via synthesized examples.
\newblock In {\em CVPR}, 2018.

\bibitem[\protect\citeauthoryear{Lampert \bgroup \em et al.\egroup
  }{2014}]{lampert2014attribute}
Christoph~H Lampert, Hannes Nickisch, and Stefan Harmeling.
\newblock Attribute-based classification for zero-shot visual object
  categorization.
\newblock {\em IEEE TPAMI}, 36(3):453--465, 2014.

\bibitem[\protect\citeauthoryear{Lankford}{1969}]{lankford1969regionalization}
Philip~M Lankford.
\newblock Regionalization: theory and alternative algorithms.
\newblock {\em Geographical Analysis}, 1(2):196--212, 1969.

\bibitem[\protect\citeauthoryear{LeCun \bgroup \em et al.\egroup
  }{2015}]{lecun2015deep}
Yann LeCun, Yoshua Bengio, and Geoffrey Hinton.
\newblock Deep learning.
\newblock {\em Nature}, 521(7553):436, 2015.

\bibitem[\protect\citeauthoryear{Liu \bgroup \em et al.\egroup
  }{2018}]{liu2018zero}
Yang Liu, Quanxue Gao, Jin Li, Jungong Han, and Ling Shao.
\newblock Zero shot learning via low-rank embedded semantic autoencoder.
\newblock In {\em IJCAI}, 2018.

\bibitem[\protect\citeauthoryear{Mahajan \bgroup \em et al.\egroup
  }{2018}]{mahajan2018exploring}
Dhruv Mahajan, Ross Girshick, Vignesh Ramanathan, Kaiming He, et~al.
\newblock Exploring the limits of weakly supervised pretraining.
\newblock In {\em ECCV}, 2018.

\bibitem[\protect\citeauthoryear{Nickel and Kiela}{2017}]{nickel2017poincare}
Maximillian Nickel and Douwe Kiela.
\newblock Poincar{\'e} embeddings for learning hierarchical representations.
\newblock In {\em NIPS}, 2017.

\bibitem[\protect\citeauthoryear{Nitta and Kuroe}{2018}]{nitta2018hyperbolic}
Tohru Nitta and Yasuaki Kuroe.
\newblock Hyperbolic gradient operator and hyperbolic back-propagation learning
  algorithms.
\newblock {\em IEEE TNNLS}, 29(5):1689--1702, 2018.

\bibitem[\protect\citeauthoryear{Parikh and Grauman}{2011}]{parikh2011relative}
Devi Parikh and Kristen Grauman.
\newblock Relative attributes.
\newblock In {\em ICCV}, 2011.

\bibitem[\protect\citeauthoryear{Patterson and Hays}{2012}]{patterson2012sun}
Genevieve Patterson and James Hays.
\newblock Sun attribute database: Discovering, annotating, and recognizing
  scene attributes.
\newblock In {\em CVPR}, 2012.

\bibitem[\protect\citeauthoryear{Roy \bgroup \em et al.\egroup
  }{2018}]{roy2018unsupervised}
Debaditya Roy, Sri Rama~Murty Kodukula, et~al.
\newblock Unsupervised universal attribute modelling for action recognition.
\newblock {\em IEEE TMM}, 2018.

\bibitem[\protect\citeauthoryear{Shepard}{1968}]{shepard1968two}
Donald Shepard.
\newblock A two-dimensional interpolation function for irregularly-spaced data.
\newblock In {\em ACM National Conference}, 1968.

\bibitem[\protect\citeauthoryear{Tay \bgroup \em et al.\egroup
  }{2018}]{tay2018hyperbolic}
Yi~Tay, Luu~Anh Tuan, and Siu~Cheung Hui.
\newblock Hyperbolic representation learning for fast and efficient neural
  question answering.
\newblock In {\em ACM WSDM}, 2018.

\bibitem[\protect\citeauthoryear{Toussaint}{1980}]{toussaint1980relative}
Godfried~T Toussaint.
\newblock The relative neighbourhood graph of a finite planar set.
\newblock {\em Pattern Recognition}, 12(4):261--268, 1980.

\bibitem[\protect\citeauthoryear{Welinder \bgroup \em et al.\egroup
  }{2011}]{welinder2011caltech}
Peter Welinder, Steve Branson, Takeshi Mita, Catherine Wah, et~al.
\newblock The caltech-ucsd birds-200-2011 dataset.
\newblock Technical Report CNS-TR-2011-001, California Institute of Technology,
  2011.

\bibitem[\protect\citeauthoryear{Xian \bgroup \em et al.\egroup
  }{2018a}]{xian2018zero}
Yongqin Xian, Christoph~H Lampert, Bernt Schiele, and Zeynep Akata.
\newblock Zero-shot learning-a comprehensive evaluation of the good, the bad
  and the ugly.
\newblock {\em IEEE TPAMI}, 2018.

\bibitem[\protect\citeauthoryear{Xian \bgroup \em et al.\egroup
  }{2018b}]{xian2018feature}
Yongqin Xian, Tobias Lorenz, Bernt Schiele, and Zeynep Akata.
\newblock Feature generating networks for zero-shot learning.
\newblock In {\em CVPR}, 2018.

\bibitem[\protect\citeauthoryear{Xu \bgroup \em et al.\egroup
  }{2017}]{xu2017matrix}
Xing Xu, Fumin Shen, Yang Yang, Dongxiang Zhang, Heng~Tao Shen, and Jingkuan
  Song.
\newblock Matrix tri-factorization with manifold regularizations for zero-shot
  learning.
\newblock In {\em CVPR}, 2017.

\bibitem[\protect\citeauthoryear{Zhang \bgroup \em et al.\egroup
  }{2018a}]{zhang2018kernel}
Zhao Zhang, Lei Jia, Mingbo Zhao, Guangcan Liu, Meng Wang, and Shuicheng Yan.
\newblock Kernel-induced label propagation by mapping for semi-supervised
  classification.
\newblock {\em IEEE Transactions on Big Data}, 2018.

\bibitem[\protect\citeauthoryear{Zhang \bgroup \em et al.\egroup
  }{2018b}]{zhang2018robust}
Zhao Zhang, Fanzhang Li, Lei Jia, Jie Qin, Li~Zhang, and Shuicheng Yan.
\newblock Robust adaptive embedded label propagation with weight learning for
  inductive classification.
\newblock {\em IEEE TNNLS}, 29(8):3388--3403, 2018.

\bibitem[\protect\citeauthoryear{Zhou}{2017}]{zhou2017brief}
Zhi-Hua Zhou.
\newblock A brief introduction to weakly supervised learning.
\newblock {\em National Science Review}, 5(1):44--53, 2017.

\bibitem[\protect\citeauthoryear{Zoidi \bgroup \em et al.\egroup
  }{2018}]{zoidi2018positive}
Olga Zoidi, Anastasios Tefas, Nikos Nikolaidis, and Ioannis Pitas.
\newblock Positive and negative label propagations.
\newblock {\em IEEE Trans. Circuits Syst. Video Technol.}, 28(2):342--355,
  2018.

\end{thebibliography}

\end{document}